\documentclass[conference]{IEEEtran}
\pdfoutput=1
\usepackage{cite}
\usepackage{amsmath,amssymb,amsfonts}

\usepackage{makecell}
\usepackage{stfloats}
\usepackage{cite}
\usepackage{dsfont}
\usepackage{balance}
\usepackage{diagbox}
\usepackage{algorithmic}
\usepackage{algorithm}
\usepackage{graphicx}
\graphicspath{ {./images/} }
\usepackage{textcomp}
\usepackage[dvipsnames]{xcolor}

\usepackage{multicol}
\usepackage{xcolor}
\usepackage{makecell}
\usepackage{stfloats}
\usepackage{cite}
\usepackage{dsfont}
\usepackage{adjustbox}
\usepackage{multirow}
\usepackage{relsize}
\usepackage{enumitem}

\makeatletter
\def\ps@IEEEtitlepagestyle{%
  \def\@oddfoot{\mycopyrightnotice}%
  \def\@evenfoot{}%
}
\def\mycopyrightnotice{%
  {\footnotesize 978-1-6654-4331-9/21/\$31.00 \textcopyright2022 IEEE
\hfill}
  \gdef\mycopyrightnotice{}
}

\newcolumntype{J}{>{\raggedright\arraybackslash}m{2.7cm}}     
\newcolumntype{K}{>{\centering\arraybackslash}m{0.9cm}}    
\newcolumntype{L}{>{\raggedright\arraybackslash}m{3.7cm}}
\newcolumntype{M}{>{\raggedright\arraybackslash}m{1.4cm}}
\newcolumntype{N}{>{\RaggedRight\hspace{0pt}}X}

\makeatletter

\let\old@ps@headings\ps@headings
\let\old@ps@IEEEtitlepagestyle\ps@IEEEtitlepagestyle
\def\confheader#1{%
    \def\ps@IEEEtitlepagestyle{%
        \old@ps@IEEEtitlepagestyle%
        \def\@oddhead{\strut\hfill#1\hfill\strut}%
        \def\@evenhead{\strut\hfill#1\hfill\strut}%
    }%
    \ps@headings%
}
\makeatother

\confheader{%
        \parbox{\textwidth}{\centering This article has been accepted for publication in the 41st IEEE International Performance Computing and Communications Conference (IPCCC), 2022.}
}

\usepackage[pscoord]{eso-pic}
\newcommand{\placetextbox}[3]{
\setbox0=\hbox{#3}
\AddToShipoutPictureFG{ \put(\LenToUnit{#1\paperwidth},\LenToUnit{#2\paperheight}){\vtop{{\null}\makebox[0pt][c]{#3}}}
}
}
\placetextbox{.5}{0.050}{\footnotesize {\copyright 2022 IEEE. Personal use of this material is permitted. Permission from IEEE must be obtained for all other uses, in any current or future media} }
\placetextbox{.5}{0.040}{\footnotesize { including reprinting/republishing this material for advertising or promotional purposes, creating new  collective works, for resale}} 
\placetextbox{.5}{0.030}{\footnotesize { or redistribution to servers or lists, or reuse of any copyrighted component of this work in other works.}
}

\begin{document}

\title{Digital Twin in Safety-Critical Robotics Applications: Opportunities and Challenges}

\author{\IEEEauthorblockN{
\large Sabur Baidya\IEEEauthorrefmark{1}
, Sumit K. Das\IEEEauthorrefmark{2}
, Mohammad Helal Uddin\IEEEauthorrefmark{1}, Chase Kosek\IEEEauthorrefmark{1}, Chris Summers\IEEEauthorrefmark{1}\\
\IEEEauthorblockA{\IEEEauthorrefmark{1}\normalsize Department of Computer Science and Engineering, University of Louisville, KY, USA}
\IEEEauthorblockA{\IEEEauthorrefmark{2}\normalsize Department of Electrical and Computer Engineering, University of Louisville, KY, USA}
\normalsize {e-mail:  sabur.baidya@louisville.edu, sumitkumar.das@louisville.edu, mohammad.helaluddin@louisville.edu}
}}

\maketitle


\vspace{-8mm}
\begin{abstract}
Digital Twin technology is being envisioned to be an integral part of the industrial evolution in modern generation. With the rapid advancement in the Internet-of-Things (IoT) technology and increasing trend of automation,  integration between the virtual and the physical world is now realizable to produce practical digital twins. However, the existing definitions of digital twin is incomplete and sometimes ambiguous. Herein, we conduct historical review and analyze the modern generic view of digital twin to create its new extended definition. We also review and discuss the existing work in digital twin in safety-critical robotics applications. 
Especially, the usage of digital twin in industrial applications necessitates autonomous and remote operations due to environmental challenges. However, the uncertainties in the environment may need close monitoring and quick adaptation of the robots which need to be safety-proof and cost effective. We demonstrate a case study on developing a framework for safety-critical robotic arm applications and present the system performance to show its advantages, and discuss the challenges and scopes ahead.

\end{abstract}

\begin{IEEEkeywords}
 Digital Twin, Robotics, Internet of Things, Industrial IoT, Robotic Arm, Robot Operating System (ROS) 
\end{IEEEkeywords}

\vspace{-4mm}
\section{Introduction}
Digital Twin as a concept has garnered significant attention over the last few years and considered as one of the top strategic technology trends in the last five years~\cite{sharma2020digital}
. This can be seen by the estimated market of digital twin applications to exceed 35 billion USD by the year 2025~\cite{jiang2021digital}.
This attention is only increasing as the industrial automation is pushing towards more and more integration between the virtual and the physical world. With the advancement of the Internet-of-Things (IoT) technologies~\cite{li2015internet}, the modern smart sensors, actuators, computing and communication devices have enabled the seamless collection and exchange of information between the physical and virtual space, thus making the practical implementation of the digital twin possible. Furthermore, the rapid progress in  artificial intelligence (AI), data analytics and edge/cloud computing has facilitated the support needed for critical applications with stringent performance constraints. 

Digital Twin technology is revolutionizing many important fields including the manufacturing industry, connected and autonomous vehicles, healthcare IoT, construction, city planning and many others~\cite{qi2019enabling}. With the advancement of robotics and automation, most of these  applications involve infrastructure based on robotics with varying degrees of autonomy and human cooperation. Although the application of robotics in these domains aims to increase the performance and support tasks that are difficult and/or impossible for human, a large factor for using robotics is enabling scalability and  minimizing the cost of operation. Additionally, the safety-critical robots in industrial applications necessitates autonomous and remote operations due to environmental challenges. However, the uncertainties in the environment needs close monitoring and quick adaptation of the robots to ensure safety. Recent works proposed learning based adaptation of the 
robot~\cite{matulis2021robot} which however is not full safety-proof in uncertain and dynamic environments, e.g., in sensitive applications, a single error can have domino effect damaging the industrial process. To mitigate the problem, digital twin based framework is proposed where a virtual replica of the real robots will be continuously monitored in real-time. This is in lieu of complete tele-operation of the robot, where autonomy is incorporated in the physical world, and only in case of anomaly in the behaviour of robot or environment, human intervention or cooperation is needed, thus saving the operational cost significantly.

Herein,  we first present a review of digital twin to create a new extended definition and conduct a survey of digital twin in the  robotics applications. Then we show a case study on a digital twin of a robotic arm application based on Franka-Emika-Panda robot, 
and show the performance of the bidirectional digital twin with a few scenarios. 

 

\vspace{-2mm}
\section{A Primer on Digital Twin}
The digital twin conceptual model was first introduced in 2003 in a presentation about product life-cycle management by Michael Grieves. Though the term was coined in the early 2000's the first actual usage of a digital twin was several years later by The National Aeronautics and Space Administration (NASA) where a digital twin was implemented in order to mirror conditions in space for testing and flight preparations for the actual physical hardware~\cite{sharma2020digital, fuller2020digital}. NASA also utilized Digital Twins in order to better predict the long term performance of space and air crafts~\cite{HAAG201864}. 
Since then, over the following years, the rapid advancement in IoT, cloud computing,  and big data analytics,  along with better 
means to collect information had become more automated, the simulation of real world physics had greatly advanced, and real-time processing of big data has made the digital twin to fruition \cite{wang2020digital}.

\begin{figure*}[!t]
\centering
 \includegraphics[width=0.8\linewidth]{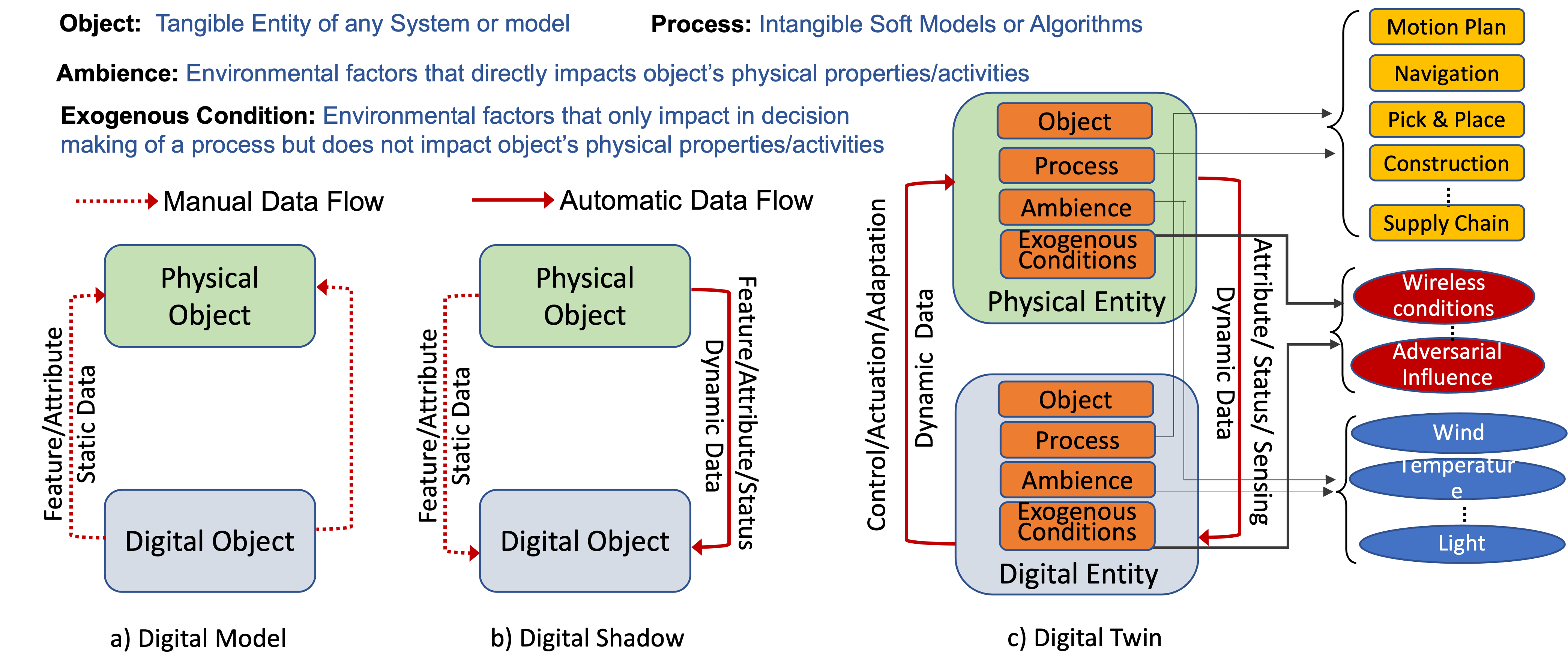}
 \vspace{-4mm}
 \caption{Comparison of Digital Twin with Digital Model and Digital Shadow}
 \label{fig:miscons}
  \vspace{-6mm}
\end{figure*}

\vspace{2mm}
\noindent \emph {\textbf {So what exactly is a Digital Twin?}} \\
There are many definitions of digital twin, and there are a lot of misconceptions as well. One of the popular definition comes from the first use case by NASA which defines it as an ``{\it integrated multiphysics, multiscale, probabilistic simulation of an as-built vehicle or system that uses the best available physical models, sensor updates, fleet history, etc., to mirror the life of its corresponding flying twin}." \cite{glaessgen2012digital}. This definition is specific to a Digital Twin of an aviation and space sense. However, the basics remain -  {\it a Digital Twin is a software model of a physical device which allows for the simulation of different environments, time periods, and variables in a device's lifespan while also actively accounting for sensor data and operational changes}\cite{grieves2017digital}\cite{sharma2020digital}. 

\par
The snowball effect of digital twin's popularity continued to rise in 2013 as `Industry 4.0'~\cite{ghobakhloo2020industry} is proposed utilizing the digital twin model. 
The combination of digital twins and the Industrial IoT (IIoT) can help realize the goal of Industry 4.0.


\vspace{-2mm}
\subsection*{\bf Definition Ambiguity:}
\vspace{-1mm}
 The ambiguous definitions and usage of the term ``Digital Twin" in various applications sometimes lead to confusion between a digital twin and similar sounding computer based simulations
, e.g., `Digital Models' and `Digital Shadows'.
A digital model usually creates a 3D or 2D construction of a conceptual object in a digital environment for efficient design of that object in real-world. 
Here only static data related to feature and attributes of the object flows manually from the digital to the physical model as shown in fig.~\ref{fig:miscons}(a). So in a Digital Model, any physical changes to the real-world model have no impact whatsoever on the digital equivalent \cite{fuller2020digital}. An example of digital model would be using AutoCAD software building schematics or a device design plans where the real-world object, or device does not have any impact on its digital  model. 
This is a big difference with respect to a digital twin, which not only does future predictions in the physical object based on the design with the digital counterpart but, also report the real-world changes in the physical entity to the digital counterpart through sensor  data feedback into the software \cite{fuller2020digital} for adaptation plan in the physical world.

 \par
 The Second common misconception is that a digital twin is same as a `Digital Shadow'. In a digital shadow, the digital replica of a physical object is taken with scan or reflection~\cite{sepasgozar2018evaluation}. Here, data flow can be manual in case of static information, or automatic for receiving dynamic scanned information. 
 But it is unidirectional from the physical object to the digital counterpart.
 This means that the digital object can successfully receive sensor data and changes in the state of the physical model but, cannot influence the physical model itself as shown in fig.~\ref{fig:miscons}(b). So an example of this would be using data from a sensor to monitor industrial machines or vehicles but not change or simulate their states in real-world.
 
 \par
 The digital twin, however, establishes a bidirectional connection between the physical and virtual entity as shown in fig.~\ref{fig:miscons}(c). Similar to digital model and shadow, one of the components in these entities is the object which can be a tangible machine or system and their physics and kinematics behavior. Additionally, the entities can  consist of processes implemented with software algorithms, e.g., motion plan, pick and place task, construction, supply chain etc. Moreover, the entities can contain the  environmental characteristics, e.g., wind, temperature, light that may impact the physical properties or activities of the object or the process. An entity can also contain exogenous conditions which do not directly impact or interact with the physical properties or activities of the object/process, but can impact the decision making, e.g., the wireless channel conditions or any adversarial influence as shown in fig.~\ref{fig:miscons}(c). For example, in a collaborative multi-robot task, the physical and digital twin need to produce the replica of the robots, their physics, algorithm for the task, any adverse weather conditions as it can impact the robot's movements, and wireless conditions which is a function of indoor structures, their materials, any presence of wireless interference - as they all can impact the collaboration of robots over wireless communications. Any dynamic changes in the object, process or environment can be detected in real-time with sensors and sent to the digital entity. Similarly, once the detected changes can be faithfully reproduced at the digital twin and tested for possible safe adaptations of the system, the updated controls can be sent to the physical twin.
 

\noindent \emph {\textbf{Extended Definition of Digital Twin:}}

Based on the aforementioned analysis and discussions, we aim to create an extended definition of digital twin as follows:\\

\vspace{-3mm}
{\it A digital twin  framework involves a `physical entity' consisting of objects, processes, interacting ambience and exogenous conditions, which are digitally reproduced in a counterpart `digital entity', and a bidirectional information flow between the physical and digital entity ensures the state and control information exchanges between them, supporting synchronous or asynchronous behavioral influence on each other.}

\vspace{1mm}
Note that, the bidirectional information flow can be  real-time/synchronous for dynamic adaptation of the system in mission-critical applications, or  can be non-real-time/asynchronous for monitoring the state of the system with the sensor data and take action later. The definition does not include specific technologies as they are the enablers of digital twin implementation and not the fundamental part of the concept.

\nocite{choi2022integrated, li2022ar, matulis2021robot, garg2021digital, fukushima2021digimobot, havard2019digital, droder2018machine, chhetri2019quilt, kousi2019digital, el2018digital, wang2020digitalweld, liu2020digital, laaki2019prototyping, mo2021terra, rassolkin2019digital, wang2020digitalv, venkatesan2019health}

\begin{table*}[!t]
  \centering
  \includegraphics[angle=-90, width=0.95\linewidth]{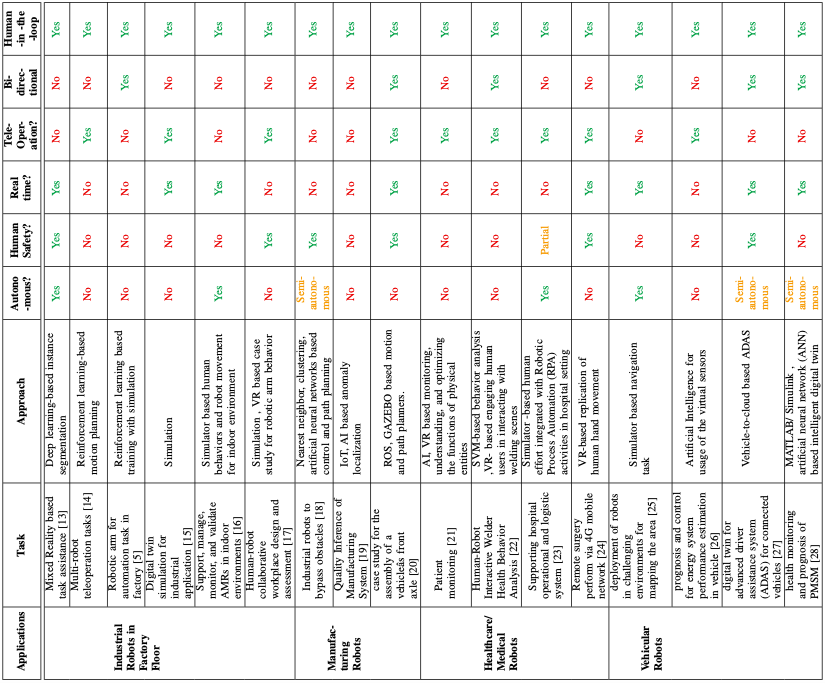}
  \caption{Digital Twin in robotics in various application domains}
  \vspace{-6mm}
  \label{tab:xxx}
\end{table*}

\section{Application of Digital Twin in Robotics}
In this paper, we focus only on the robotics applications in different domains and discuss how digital twin is proposed to be used for those applications.

\subsection{Manufacturing and Factory Floor}
In industrial robotics applications, a digital twin's ability to simulate the physical system allows real-time synchronization and decision making which can be used to choose the optimized actions for improved efficiency, accuracy, and economic gains in the production \cite{KRITZINGER20181016, matulis2021robot, garg2021digital}. It can help inferring the quality of a manufacturing process as well~\cite{chhetri2019quilt, kousi2019digital}. 
The twins have the ability to identify how changes affect the upstream and downstream processes. This allows for better scheduling with increased efficiency. 
A digital twin for the industrial robots can help factories to better understand their machines health conditions and needs, which can increase competitiveness, productivity, and efficiency \cite{KRITZINGER20181016}\cite{he2018surveillance}.
This gives a lot more power in predictive measures and in analyzing data and how machines work\cite{kamath2020industrial}\cite{augustine2020industry}. Table I summarizes some of the industrial robotic applications along with indications of the operational and safety characteristics of the robots considered in the respective applications.

\subsection{Healthcare}
\par
Application of robotics in healthcare and medicine has rapidly increased since the COVID-19~\cite{zemmar2020rise}.
The digital twin can play a vital role in reducing the cost of treatment for the patient. Accurate digital twin modeling can provide great insight on deteriorating conditions and allow for better tailored treatment plans. These advanced diagnostic tools can be used to supplement other systems already in place. They can also help provide better healthcare in developing countries~\cite{augustine2020industry}.


\par

\par
One of the most popular healthcare application of digital twin is automated remote health monitoring of patients incorporating various sensors, AR/VR technologies and AI-driven algorithms~\cite{el2018digital,wang2020digitalweld}. Another cutting-edge medical applications where robotics is envisioned to be greatly useful, is the remote surgery. A digital twin based remote surgery~\cite{laaki2019prototyping} can increase the accuracy, safety and efficiency of the medical surgery. Digital twin can also be used to improve the hospital's operational and logistic system~\cite{liu2020digital}. All these proposed digital twin models satisfy varying degrees of operational and safety needs of the healthcare robotic systems as shown in Table I.

\subsection{Connected and Autonomous Vehicles}
Through coupling vehicles with their digital twins, the automobile manufacturers can retrieve functional information on the vehicle and help examining of performance of vehicles through their lifetimes and can help in suggestive and preventative maintenance of their automobiles~\cite{rassolkin2019digital, venkatesan2019health, augustine2020industry}.
Another functionality of digital twin in vehicular application, is testing the vehicles in complex simulated environments and scenarios before deploying the physical model in real-world. 
For similar reasons NASA and the Air Force adopted digital twin to test with lower cost the conditions that may be impossible to replicate easily
\cite{wang2021digital}. Digital twin for vehicles is also used to map the challenging environments~\cite{mo2021terra} and for advanced driving assistance system (ADAS)~\cite{wang2020digitalv}.
In table I, we summarize the digital twin propositions for all these robotics applications and compare them with respect to their characteristics and performance offerings.

\section{Case Study: A Digital Twin for Safety-Critical Robotic Arm in Uncertain, Dynamic Environment}

The digital twin of a robotic arm is used for different applications in the industry floor where the robot can perform various heavy-duty tasks and the status of the robot as well as their functionality can be monitored over a digital twin.  
However, the robots may often face uncertainties due to dynamics in the environment or anomaly introduced by an adversary. Overcoming these uncertainties may need human intervention or re-training the autonomous decision-making models to update the motion plan. In safety-critical applications, autonomous re-planning of the motion and deploying directly on the physical robot is not fully safety-proof. Instead capturing the dynamic uncertainty and reproducing on the digital counterpart to retrain the motion planning algorithm and testing on the digital model before deploying on the physical model is ideal.


\begin{figure}[!t]
\centering
\includegraphics[width=0.90\linewidth]{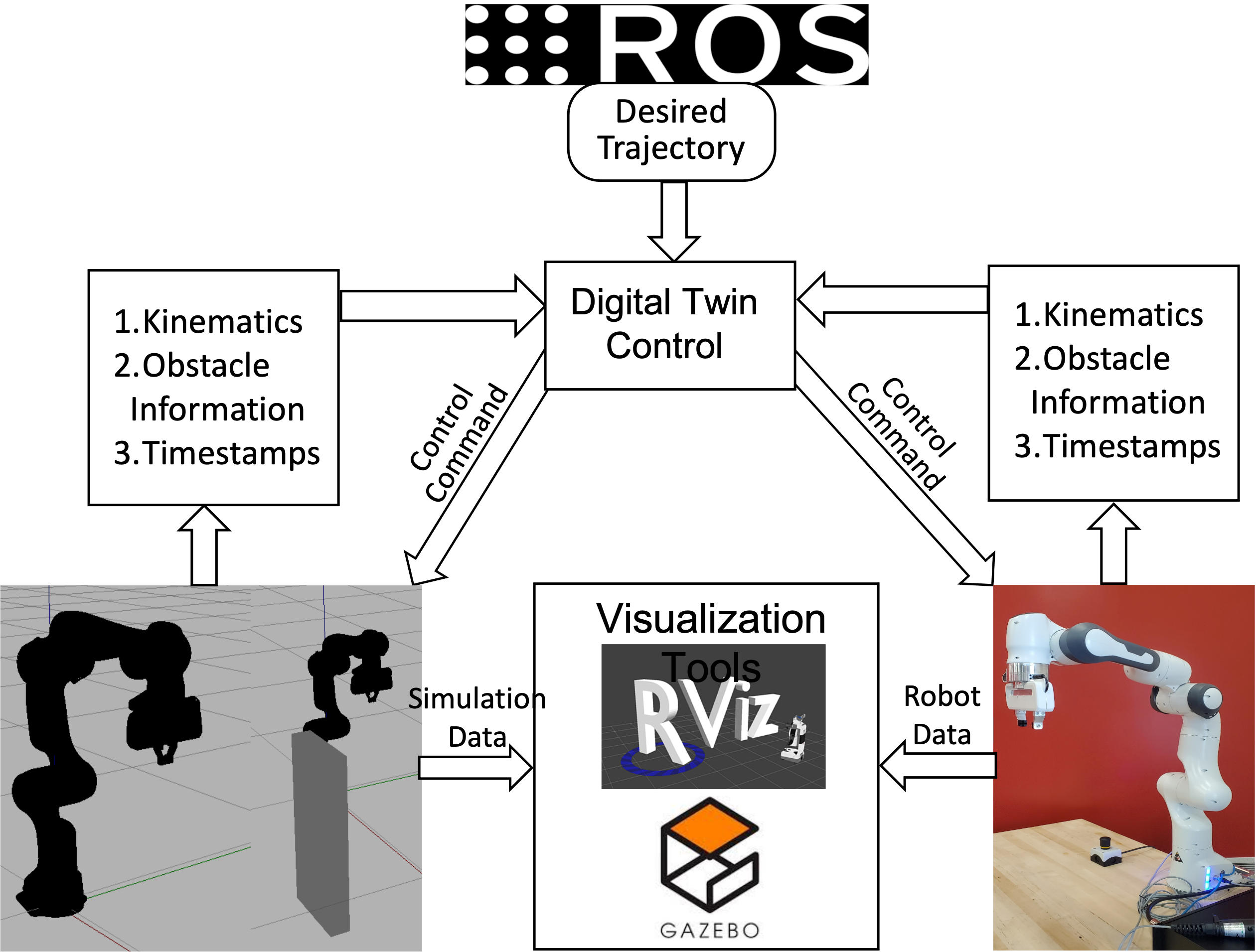}
\vspace{-2mm}
\caption{ROS-based Bi-Directional Information Architecture of Franka Robotic Arm Digital Twin}
\label {fig:twinControl}
\end{figure}

\setlength{\textfloatsep}{0pt}
\begin{algorithm} [!t]
\smaller
 \caption{Bi-Directional Digital Twin Control}
 \begin{algorithmic}[1]
 \label{algo:twinControl}
 \renewcommand{\algorithmicrequire}{\textbf{Input:}}
 \renewcommand{\algorithmicensure}{\textbf{Output:}}
 \REQUIRE $P_r$, $P_v$, $TS_r$, $TS_v$, $K_r$, $K_v$, $P_{obs}$, $T_u$
 \ENSURE  $U_r$, $U_v$
 \\ \textit{Initialisation} :
  \STATE Acquire initial values for $P_r$, $P_v$, $TS_r$, $TS_v$, $K_r$, $K_v$, $P_{obs}$
 \\ \textit{LOOP Process}
  \STATE Acquire trajectory goal from user ($T_u$) 
  \WHILE {($P_r$, $P_v$) $\neq$ $T_u$}
  \STATE Send Command to move the physical robot and virtual robot by $\Delta T_u$
  \STATE Acquire values for $P_r$, $P_v$, $TS_r$, $TS_v$, $K_r$, $K_v$, $P_{obs}$
      \IF {(($P_r$, $P_v$) - $P_{obs}$) $\geq \Delta B$ }
      \STATE Alert User
      \STATE Take Obstacle Avoidance Measures
      \ENDIF
      
      \IF {$P_r - P_v \geq \Delta Q$}
      \STATE Alert User
      \STATE Record Data for Analysis
      \ENDIF
      
      \IF {$TS_r - TS_v \geq \Delta \alpha$}
      \STATE Alert User
      \STATE Record Data for Analysis
      \ENDIF
  \ENDWHILE
 \end{algorithmic} 
 \end{algorithm}

\subsection{System Perspective and Challenges}
Our aim is to analyze the system challenges to optimize the performance and the cost of digital twin operation. First challenge is that the safety-critical applications should be synchronous and blocking on the motion plan to overcome any uncertainty safely. If it is asynchronous, the robot may not handle uncertainty well and get damaged or may not perform well on subsequent activities. 
Synchronizing the physical and digital robot is challenging as both run their control algorithms on two different computers with different capacity. The digital models use robot operating system (ROS) based simulations which need real-time Linux (RT-Kernel) for comparable performance with the physical robots. 
Another challenge is the low communication need between the physical and digital counterparts. 
To overcome that high volume data transmission delay, edge computing~\cite{chowdhery2018urban} can help in partial or full computation offloading optimizing the latency and accuracy of the computing.


\vspace{3mm}
\subsection{ROS-based Digital Twin with Franka Robotic Arm}
\vspace{-2mm}
Herein, a digital twin control architecture was designed to operate a Franka Emika Panda robot and its virtual counterpart, as shown in Fig. \ref{fig:twinControl}. The control architecture uses ROS with bi-directional communication to obtain feedback and send trajectory commands to the physical and virtual robot. 
The control architecture is also capable of tracking obstacles (both virtually and physically) during the robot operation and avoid any possible collision by taking appropriate obstacle avoidance measures. As illustrated in Algorithm \ref{algo:twinControl}, the control block acquires the robot pose information $P_r, P_v$ of physical and virtual robots respectively along with their kinematics ($K_r, K_v$) and time stamp data ($TS_r, TS_v$). If any obstacle is detected either virtually or physically, the obstacle's pose, $P_{obs}$, is also relayed to the control block. The control block acquires the joint positions of the robot and through the use of the kinematic chain and ROS Kinematics Dynamic Library (KDL), the end poses, $P_r, P_v$, are calculated.

\begin{figure}[!t]
\centering
\includegraphics[width=1.0\linewidth]{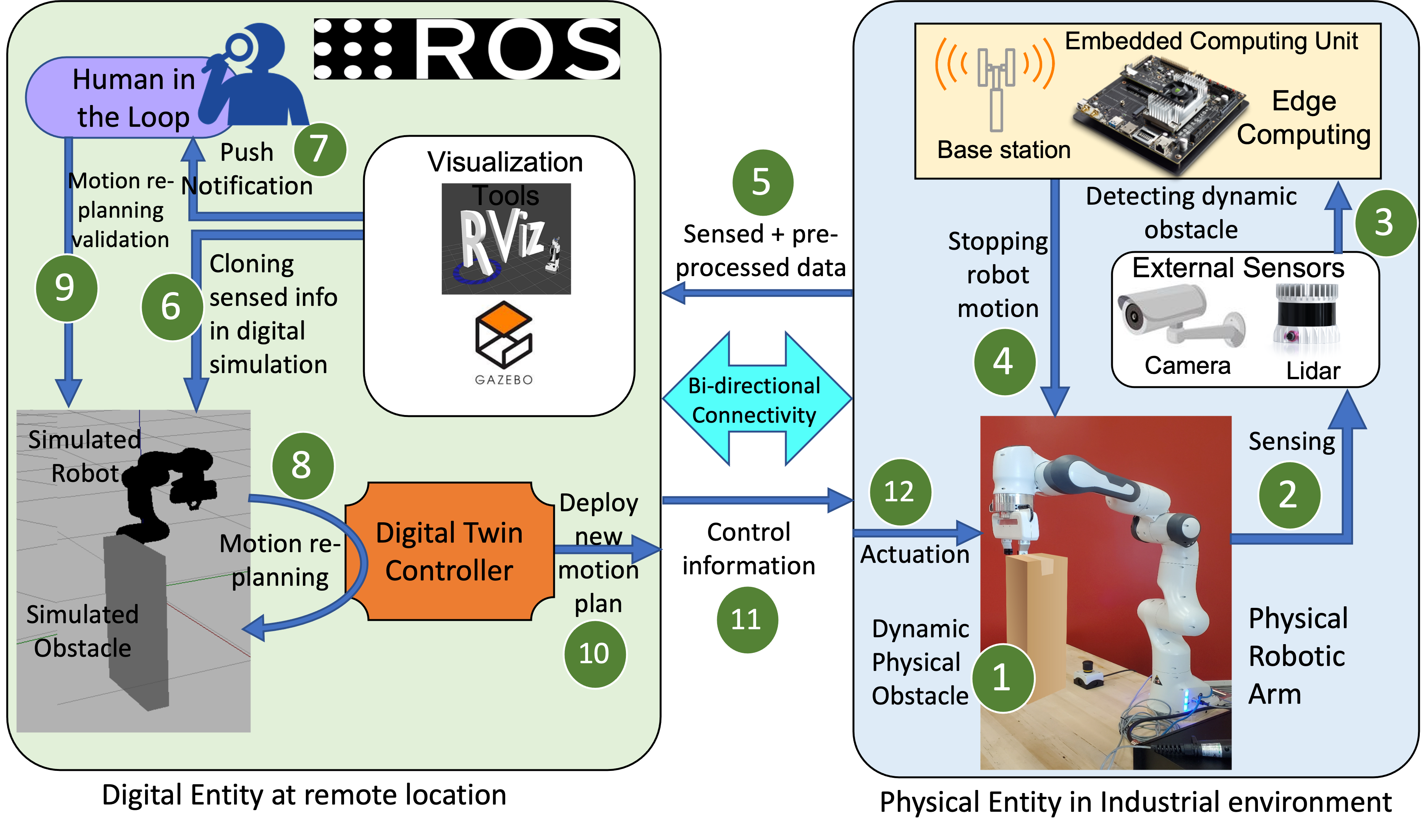}
\vspace{-8mm}
\caption{Edge-assisted Safety-proof Human-in-the-loop Framework}
\label {fig:framework}
\vspace{2mm}
\end{figure}

During the operation, the control block accepts a desired trajectory input, $T_u$, from the user and plans the path for both the physical and the virtual robot through multiple waypoints $\Delta T_u$. The control block loops through the waypoints until the desired end pose is achieved. During the movement in-between the waypoints, the control block keeps verifying that the deviation of pose and time stamp between the physical robot data and virtual robot data is not more than the set bounds of $\Delta Q$ and $\Delta \alpha$ respectively. If the deviations exceed a threshold, the system alerts the user and record the incident for further inspection. When an obstacle is detected, either virtually or physically, and the current pose of robot is with the bounds of $\Delta B$, the trajectory waypoints are modified to move the robot in a way to avoid the detected obstacle.

To make the robotic arm safety-proof in presence of dynamic obstacles or anomalies, we propose an edge-assisted human-in-the-loop framework for the digital twin of the robotic arm
as shown in Fig.~\ref{fig:framework}. It can simultaneously send a push notification to a person responsible for monitoring, 
who can check the new autonomous planning with the digital model with the dynamically introduced anomalies. Once the person approves, the new motion plan can be deployed on the physical robot which will be unblocked to use it. 






\subsection{Performance Evaluation}



\begin{figure}[!t]
\centering
\vspace{-2mm}
\includegraphics[width=0.8\linewidth]{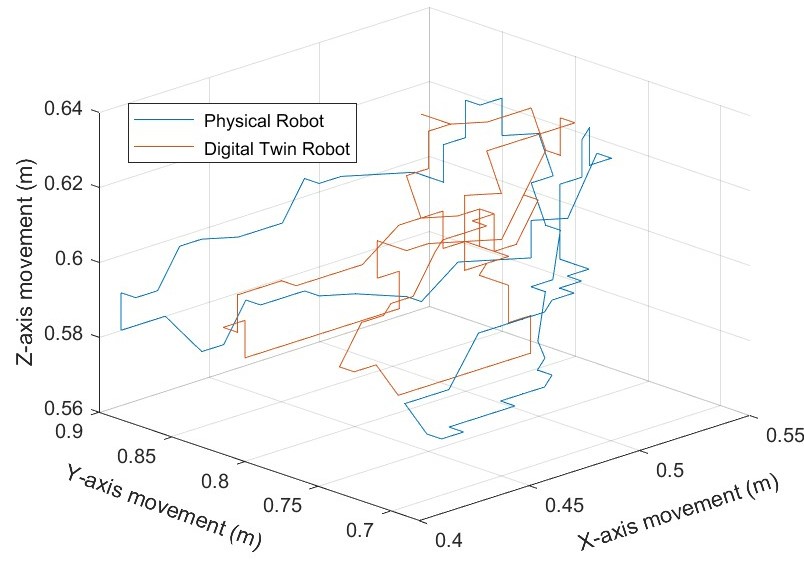}
\vspace{-3mm}
\caption{Cartesian position comparison between a physical and digital robot}
\label{fig:dt_comp}
\vspace{-4mm}
\end{figure}

\begin{figure}[!t]
\centering
\includegraphics[width=0.85\linewidth]{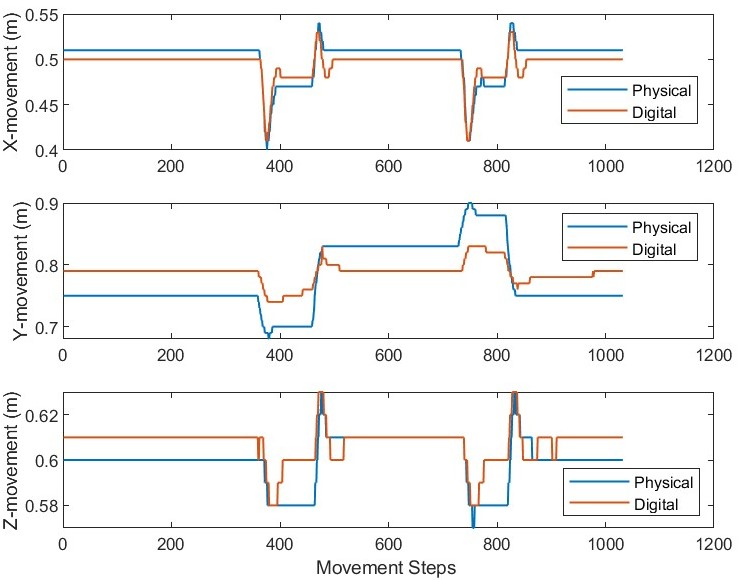}
\vspace{-3mm}
\caption{Temporal translational deviation in motion between a physical robot and digital twin robot}
\label{fig:dt_trans_err}
\end{figure}

In this case study, we implemented ROS-based digital twin with Franka-Emika robotic arm and Gazebo simulator~\cite{qian2014manipulation} for the digital robot. The ROS runs with RT-kernel on a powerful computer which runs the Gazebo simulation in almost real-time. We first performed random motion with the robotic arm  using Cartesian coordinates based waypoints and measure the movements of the end-effector in the physical and digital robots. Fig.~\ref{fig:dt_comp} shows the trajectory of both and we can see very small differences in the order of few centimeters. We also measure the errors in the motion between the physical and digital twin in individual translational and rotational movements. Fig.~\ref{fig:dt_trans_err} shows the temporal transational error 
between the robotic twins. As indicated previously, the error is quite small. Then, we measure the mean absolute error (MAE) for the experiments which shows the errors less than 5 cm as mentioned in the table II. The slight error could be created due to the difference in controller gain between the physical and digital robot. One can set up the safety margin slightly greater than this errors for seamless operation.

\begin{figure}[h]
\centering
\vspace{-2mm}
\includegraphics[width=0.8\linewidth]{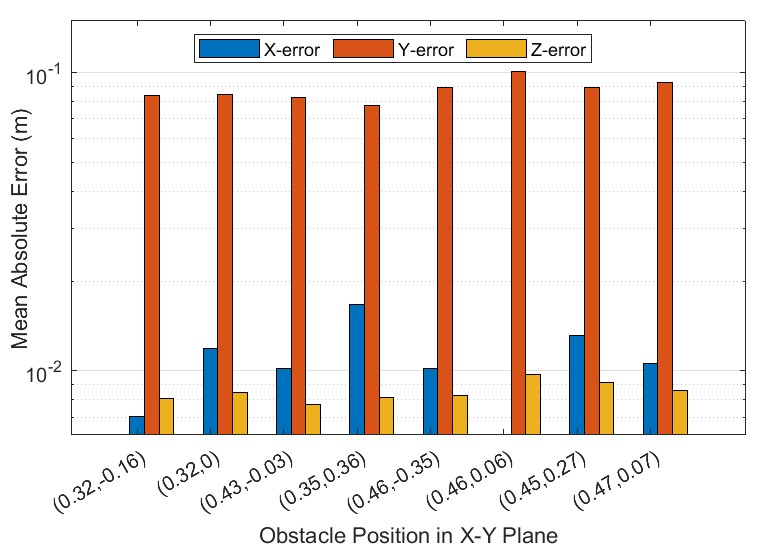}
\vspace{-3mm}
\caption{MAE of Cartesian Positions between a physical robot and digital twin robot for 0.4m high obstacle at different X-Y location}
\label{fig:mae_obs_pos}
\vspace{-3mm}
\end{figure}

\begin{table}[!b]
\centering
\vspace{2mm}
\begin{tabular}{|c|c|c|c|c|c|c|} 
 \hline
 
 {\bf X-mov} & {\bf Y-mov} & {\bf Z-mov} & {\bf Roll} & {\bf Pitch} & {\bf Yaw} & {\bf Actuation}  \\ [0.5ex] 
 {\bf (m)} & {\bf (m)} & {\bf (m)} & {\bf (rad)} & {\bf (rad)} & {\bf (rad)} & {\bf Time (ms)}\\ [0.5ex] 
 \hline
 0.0104 & 0.0401 & 0.0081 & 3.9806 & 0.0274 & 0.0401 & 16.014 \\ [0.5ex] 
 \hline
\end{tabular}
\vspace{1mm}
\caption{\small Mean Absolute Error in movements in free environment}
\label{Accu_vs_Compl}
\vspace{-3mm}
\end{table}

\begin{figure}[!t]
\centering
\includegraphics[width=0.75\linewidth]{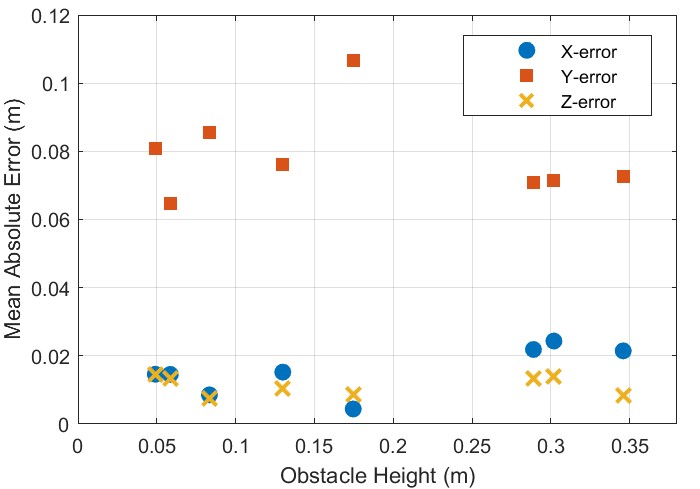}
\vspace{-3mm}
\caption{MAE of Cartesian Positions between a physical robot and digital twin robot for variable obstacle height at fixed X-Y (0.5m,0m) location}
\label{fig:mae_obs_size}
\vspace{2mm}
\end{figure}

\begin{table}[!b]
\smaller
\centering
\begin{tabular}{|J|c|c|c|c|} 
 \hline
 
 {Scenario} & {\bf X-mov} & {\bf Y-mov} & {\bf Z-mov} & {\bf Actuation}  \\ [0.5ex] 
{} & {\bf (m)} & {\bf (m)} & {\bf (m)} & {\bf Time (ms)}\\ [0.5ex] 
 \hline
 Fixed height obstacle, varying positions & 0.0107 & 0.0875 & 0.0085 & 16.013 \\
 \hline
 Fixed position obstacle, varying heights & 0.0156 & 0.0786 & 0.0113 & 16.055 \\
 [0.5ex] 
 \hline
\end{tabular}
\vspace{1mm}
\caption{\small Mean Absolute Error in Movements with dynamic varying obstacles}
\label{Accu_vs_Compl}
\vspace{-4mm}
\end{table}

We also measure the difference in actuation time in physical and digital robot when the control commands are fired simultaneously for both. Table II shows the mean absolute difference in actuation time is about 16 ms, which can't be differentiated by human eye and can be considered as real-time. 

We introduced dynamic obstacle detection and avoidance in the digital twin. For demonstration, we assumed that the obstacle is detected a priori and the information about its position and size is sent to the digital model. In the digital simulation then a simulated version of the obstacle is dynamically created and new motion plan with obstacle avoidance is executed. For consistency, we executed the same task with simple motion from left to right  along the Y-axis avoiding the obstacle. We first tested a fixed size obstacle and then randomly place at different positions to create dynamic uncertainties. Fig.~\ref{fig:mae_obs_pos} shows the mean absolute error in movements in $X, Y, Z$ directions. It can be noted that for different position of the obstacle the error does not vary proportionally. The errors in $X$ and $Z$ direction is still within 5 cm as it was in free environment. However, as the robot is moving in $Y$ direction and trying to avoid obstacle, the error slightly increase to about 10 cm. Similarly, in Fig.~\ref{fig:mae_obs_size}, we keep the location of the obstacle fixed but vary its height. It shows a similar trend as the previous case, where the errors do not vary with the obstacle height and remains within bounds. However, similar to previous case, the errors in $X$ and $Z$ direction is less than 3 cm, whereas maximum error in $Y$ direction is about a 10 cm. Table III summarizes the mean absolute error for the aforementioned cases along with the difference is actuation time which still remains about 16 ms.

\section{Conclusion and Future Scopes}
The Digital Twin creates a virtual clone with the ability to predict, monitor, share data, and control the physical twin. 
These benefits, however, are not without challenges - including data handling, communications, synchronization between the twins, simulation software, and cyber security concerns. 
Although, the cost to develop, test, and implement a digital twin is high,
the long term reward of improving system efficiency makes it worth investing.

In this paper, we revisited the fundamentals of the digital twin concept and come up with an extended definition. We 
conducted a brief survey on related works of digital twin in various robotics applications. Finally, we demonstrated a case study on the digital twin of a robotic arm and presented the performance in dynamic uncertain conditions. We proposed a safety-proof human-in-the-loop digital twin framework for safety-critical robotics applications and discuss its advantages, and challenges. In future the framework can be augmented with overlay technologies including Artificial Intelligence and Virtual and Augmented Reality (VR/AR) to create a seamless and efficient digital twin. The future robotic digital twin can leverage the advancements in 5G and upcoming 6G wireless technologies together with strong secured sensing, communications and computing to become more resilient. Also, sophisticated and ultra low latency algorithms need to be developed in future for fast-responding, robust and efficient practical digital twin in safety-critical robotics.


\section*{Acknowledgment}
This work is partially supported by the US National Science Foundation (EPSCoR \#1849213).

\balance

\bibliographystyle{IEEEtran}
\bibliography{digital-twin-bib}

\end{document}